\def\year{2017}\relax
\documentclass[letterpaper]{article} 
\usepackage{aaai17}  
\usepackage{times}  
\usepackage{helvet}  
\usepackage{courier}  
\usepackage{url}  
\usepackage{graphicx}  
\usepackage{times}

\usepackage[round]{natbib}
\usepackage{latexsym}
\usepackage[utf8]{inputenc}
\usepackage[font=small,labelfont=bf]{caption}
\usepackage{amsmath}
\usepackage{titlepic}
\usepackage{multirow}
\usepackage{appendix}
\usepackage{covington}
\usepackage{url}
\usepackage{tikz}
\usepackage{subfiles}

\usepackage{avm}
\avmfont{\sc}
\avmoptions{sorted}
\avmvalfont{\rm}
\avmsortfont{\scriptsize\it}

\usepackage{epigraph}
\usepackage{listings}
\usepackage{avm}

\avmfont{\sc}
\avmoptions{sorted}
\avmvalfont{\rm}
\avmsortfont{\scriptsize\it}
\usepackage{remreset}
\usepackage{pifont}
\usepackage{graphicx}
\usepackage{verbatim} 
\usepackage{linguex}
\usepackage{qtree}
\usepackage{algorithm}
\usepackage{algpseudocode}
\usepackage{amsmath,amsfonts,amssymb}
\usepackage{lipsum}       
\usepackage{changepage} 
\usepackage{afterpage}
\usepackage[final]{pdfpages}
\usepackage{multicol}
\usepackage{wrapfig}
\usepackage{balance}  

\algrenewcommand{\algorithmiccomment}[1]{ \# #1}

\newcommand{\wac}{\textsc{wac}}

\title{Symbol, Conversational, and Societal Grounding with a Toy Robot}

\author{Casey Kennington \& Sarah Plane\\
Department of Computer Science\\
Boise State University\\
firstnamelastname@boisestate.edu \\
}

\begin{document}
%

\maketitle
\begin{abstract}
Essential to meaningful interaction is grounding at the symbolic, conversational, and societal levels. We present ongoing work with Anki's Cozmo toy robot as a research platform where we leverage the recent words-as-classifiers model of lexical semantics in interactive reference resolution tasks for language grounding. 
\end{abstract}

\section{Introduction}

\begin{figure}[H]
  \centering
      \includegraphics[width=0.45\textwidth]{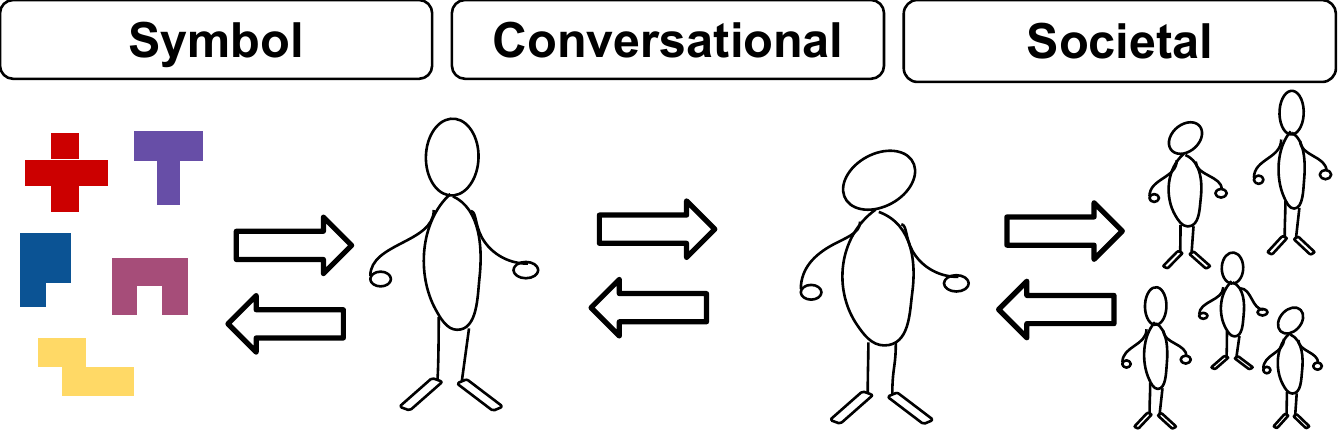}	
      \caption{\footnotesize Comparison of grounding types. An individual perceives objects and grounds symbols--conventional denotations for those objects--interactively through conversational grounding with someone else. The conventional denotations are socially grounded through interaction with members of a language community.\label{fig:grounding-types}}
\end{figure}

Grounding is essential in meaningful interaction \citep{Clark1996,devault2006societal,Schlangen2016}. Grounding is a term used to denote several distinct aspects of language and communication. We take up three aspects here, though \cite{Lucking2014} have identified others: (1) \textbf{\emph{symbol grounding}} \citep{harnard:grounding} where aspects of language are connected with aspects of the things that language denotes, such as visual features (e.g., the word \emph{red} is linked to aspects of visual perception), (2) \textbf{\emph{conversational grounding}} \citep{Clark1996} where aspects of events that occur between two or more people are recorded for later use and recall, and (3) \textbf{\emph{societal grounding}} \citep{devault2006societal} which connects symbol and conversational grounding with the accepted uses of language used in a particular language community. These aspects of grounding are summarized in \textbf{Figure~\ref{fig:grounding-types}}. 

All three types of grounding overlap with each other which allows for meaningful communication. To illustrate, consider a child who sees a pine cone and experiences first-hand its visual and tactile features. A nearby adult says ``that's a pine cone'' because the adult has already established through \emph{societal grounding} that ``pine cone'' denotes such an object. By hearing this, the child learns through \emph{symbol grounding} that certain visual and tactile features are linked to the words ``pine cone'' and both the child and adult establish through \emph{conversational grounding} the event that the child has heard the denotation. Grounding on all three levels in this example occurred through an interactive process which establishes grounding of linguistic meaning between words and the perceived world, between individuals, and between individuals and language communities at large.

It is through this face-to-face spoken conversation setting, the basic and primary setting of language \citep{Fillmore1975}, where interlocutors can denote objects (often with pointing gestures) in their shared environment which forms the foundation for language acquisition \citep{McCune2008}, and from which words denoting more abstract concepts are built. A key question is how semantic meaning should be represented and acquired through this co-located grounding process. 

We present ongoing work on grounding with a toy robot. We leverage the \emph{words-as-classifiers} (\wac) model of lexical semantics \citep{Kennington2015_acl}, recently yielding state-of-the-art results in a reference resolution task using real images and deep learning to represent the object regions \citep{Schlangenetal2016}. The model is flexible, interpretable, and simple in that each word is treated as its own classifier. 

\section{Background \& Related Work}

This work builds on related work in co-located, language grounding \citep{Roy2005} and recent work in grounded language semantic learning in various tasks and settings, notably learning descriptions of the immediate environment \citep{Walter}; navigation \citep{Kollar2010}, and verbs \citep{She}. A common task to evaluate models convincingly is reference resolution to real-world objects. In most cases, the set of candidate objects are simultaneously visible within a scene. This project goes beyond this work: the robot's limited perspective allows it to see one or two objects in front of it at a time. The robot must settle on an object potentially without being able to see all of the objects--arguably a more realistic task (similar in spirit to navigation tasks such as \cite{Hana}) and language grounding setting; i.e., the two interlocutors do not share the same perspective. Moreover, previous work has assumed that humans will treat and interact with the robot in such a way that the robot will perform symbolic grounding, but it's not necessarily the same setting where humans acquire their first language: as children. It has been shown that humans treat robots differently depending on how they perceive the robot's gender \citep{JASP:JASP937}, social categorization \citep{BJSO:BJSO2082}, personality \citep{Tay2014}, and intelligence \citep{Novikova}. In this work, we take this knowledge into account by using a robot that is more likely to be perceived and treated as a child by humans.

\begin{wrapfigure}{r}{0.15\textwidth}
  \centering
      \includegraphics[width=0.15\textwidth]{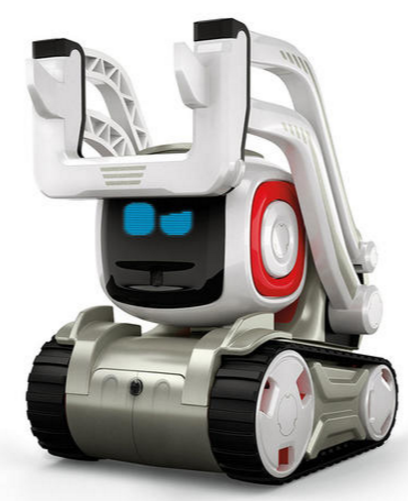}	
      \caption{\footnotesize Cozmo robot \label{fig:cozmo}}
\end{wrapfigure}

We leverage the recently released \emph{Anki Cozmo} robot as a platform to research spoken language grounding using the \wac\ model. The Cozmo robot (example in Figure~\ref{fig:cozmo}) is a small robot that has a well-documented SDK and growing community support.\footnote{https://developer.anki.com/en-us} The robot itself has arms that can lift or push small objects, track wheels for movement, a simple text to speech synthesizer (i.e., the robot itself has a small speaker), and a black and white camera which is embedded in a small movable head that has animated eyes. Some built-in capabilities include facial recognition and some basic functionality for detecting specific types of objects (e.g., some blocks that are included with the robot). The hardware that makes up the robot offer enough degrees of freedom to make it a flexible and versatile research platform; the size and affordances of the robot make it manageable for researchers who are not roboticists. The SDK is written in Python making it easily extensible by the myriad of machine learning and natural language processing libraries. Importantly, the robot is affordable (under \$180) and very portable. Our group has already acquired two Cozmo robots and we have found them to be accessible, usable, and flexible, even for fairly novice programmers.

\section{Language Grounding:\\ Our Approach}

We follow a simple strategy for language grounding and acquisition: assuming that the system can \emph{detect} (i.e., \emph{not} recognize) objects--a precondition for learning words that denote objects \cite[p.61]{bloom2000children}--we apply the \wac\ model to learn novel words with minimal interactions. We also take into account the essential pragmatic scaffolding that must be in place for language grounding to take place: the Cozmo robot can track a person's face and facial features which we will leverage for positive and negative feedback when the robot performs certain tasks that involve word usage. Learning in real-time interaction is no trivial matter, but here the Cozmo platform is useful: instead of using potentially complicated pointing recognition, we can assume that an object under discussion is the one directly in front of the robot. Evaluation of our model can be done by a reference resolution task similar to a game of fetch where a human player refers to an object and the robot must find that object as soon as possible. 

\begin{wrapfigure}{l}{0.25\textwidth}
  \centering
      \includegraphics[width=0.25\textwidth]{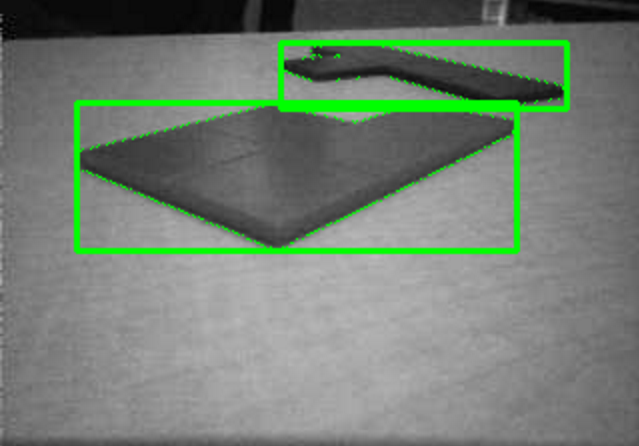}	
      \caption{\footnotesize simple object detection from Cozmo's perspective \label{fig:cozmo-view}}
\end{wrapfigure}

Our preliminary work using the Cozmo SDK has shown promise. We have applied some of our own object detection to the camera feed using OpenCV (see Figure~\ref{fig:cozmo-view}) as well as the YOLO object detection model \citep{redmon2016yolo9000}. Having detected the objects, we can extract low-level object features for the \wac\ model which does the object recognition and grounds the words in the referring expressions to the objects. In our preliminary experiments, the \wac\ model selects the correct object about half of the time with minimal training data. Supporting the \wac\ model will be additional standard dialogue system modules, such as a conversational speech recognizer and a dialogue manager. We build off of our own previous work for evaluating conversational speech recognition \citep{Baumann16} to determine the best option, and dialogue management in an interactive setting with a robot \citep{Kennington2014_hri} using the OpenDial toolkit \citep{Lison2015}. 

The outcome of this research will be improved understanding of how lexical semantic meaning is learned and represented through natural interaction. We are exploring a setting where Cozmo interacts with children to perform simple tasks, as Cozmo is marketed as a toy for children to learn procedural `coding'. In our observations, children find Cozmo aesthetically pleasing and enjoyable to interact with. We anticipate several challenges: for \wac, the robot's integrated camera has a limited, black and white perspective (i.e., the \wac\ model cannot make direct use of color information in this setting). Verb learning of robot actions will also be challenging (e.g., \emph{move}, \emph{pick up}, \emph{push}, etc.); we will build off of very recent work by \cite{She2017}. The task and setting will also challenge the \wac\ model due to differences in perspectives (e.g., the word \emph{left} will mean something different depending on the perspective of the users and the robot). 

Though we are not roboticists, we feel it important to bring together dialogue systems and robotics researchers to work towards natural, spoken interaction with robots. 

\newpage
{
\bibliographystyle{aaai}
\balance
\bibliography{refs2}
}

\end{document}